\pdfoutput=1

\documentclass[a4paper]{article}

\usepackage{INTERSPEECH2021}
\usepackage{enumitem}
\usepackage{latexsym}
\usepackage[utf8]{inputenc}

\title{Alternate Endings: Improving Prosody for Incremental Neural TTS with Predicted Future Text Input}
\name{Brooke Stephenson$^{1,2}$, Thomas Hueber${^1}$, Laurent Girin$
{^1}$, Laurent Besacier$^{2,3}$}
\address{
  $^1$Université Grenoble Alpes, CNRS, Grenoble INP, GIPSA-lab, 38000 Grenoble, France\\
  $^2$LIG, UGA, G-INP, CNRS, INRIA, Grenoble, France\\
  $^3$NAVER LABS Europe, France}
\email{brooke.stephenson@gipsa-lab.grenoble-inp.fr, 
thomas.hueber@gipsa-lab.grenoble-inp.fr,
laurent.girin@gipsa-lab.grenoble-inp.fr,
laurent.besacier@univ-grenoble-alpes.fr}

\begin{document}

\maketitle

\begin{abstract}

Inferring the prosody of a word in text-to-speech synthesis requires information about its surrounding context. In incremental text-to-speech synthesis, where the synthesizer produces an output before it has access to the complete input, the full context is often unknown which can result in a loss of naturalness. 
In this paper, we investigate whether the use of predicted future text from a transformer language model can attenuate this loss in a neural TTS system. We compare several test conditions of next future word: (a) unknown (zero-word), (b) language model predicted, (c) randomly predicted and (d) ground-truth. We measure the prosodic features (pitch, energy and duration) and find that predicted text provides significant improvements over a zero-word lookahead, but only slight gains over random-word lookahead. We confirm these results with a perceptive test.
\end{abstract}


\section{Introduction}
In incremental text-to-speech synthesis (iTTS), the system starts to output chunks of synthetic audio before the full text input is known \cite{baumann_evaluating_2012, baumann2014partial, pouget2015hmm, pouget_2016_interspeech_adaptive_latency_iTTS}. The missing input information often hinders the ability to produce a natural sounding speech sequence, mostly because prosodic features that will be determined by the future context (i.e. the remaining words in the sentence) have not yet been specified. Fortunately, the future input is not completely random; human language is characterized by several lexical and syntactic patterns, which can be statistically learnt and then predicted to a certain extent. Recent advances in language modelling, namely the use of transformer models such as BERT \cite{devlin-etal-2019-bert} and GPT-2 \cite{radford2019language} give us accurate representations of the probability distribution of future words. If this information can be mobilized to fill in the missing data for an iTTS system, it may be possible to retain naturalness while minimizing latency.

Early work in iTTS was conducted in the context of HMM-based models, where linguistic and phonological features extracted from text were used to estimate speech parameters. \cite{6854316} studied the effects of missing future features by replacing decision tree split criteria with default values and evaluating the degradation in speech quality: while cepstral and aperiodicity features could be estimated fairly accurately with just a local context, prosodic features (f0 and duration) were found to be more dependent on longer range context. Using a similar default value strategy, \cite{baumann2014partial} studied symbolic intonation assignment in the presence/absence of word, phrase and utterance level features. They report that phrase and utterance final words benefit the most from the inclusion of phrase and utterance level features in prosodic assignment determination.  \cite{pouget2015hmm} explicitly specified unknown features in the context clustering process and found improvements over a default value strategy.
\cite{baumann_evaluating_2012} tested just-in-time strategies for integrating future chunks in a dialogue system and concluded that incorporating the next phrase once the first word of the current phrase had been processed gave the best latency/quality trade-off. Finally \cite{pouget_2016_interspeech_adaptive_latency_iTTS} developed an adaptive policy which delayed synthesis when there was high uncertainty regarding POS tags.  

Recent research in iTTS has focused on end-to-end neural models. While these models create more natural speech, they are also more difficult to analyze because the relevant features for the task are learnt during training and are subsequently not easily human interpretable. In this new paradigm, the trade-off between speech quality and synthesis latency has been examined by testing the effects of different degrees of lookahead \cite{ stephenson2020future, ma-etal-2020-incremental}, reinforcement learning has been used to automatically learn a wait/synthesize policy \cite{Mohan2020IncrementalTT}, the effects of synthesis unit have been studied \cite{yanagita2019neural} and the integration of iTTS into a speech-to-speech translation system \cite{sudoh2020simultaneous} and into a machine speech chain \cite{Novitasari2020IncrementalMS} have been evaluated.  

Furthermore, previous studies in conventional (i.e.~non incremental) TTS have incorporated language model representations into neural TTS training and found that they could help speed up training time \cite{fang2019towards, chung2019semi}. 
In the field of simultaneous translation, where future context is also unknown, \cite{zheng-etal-2019-speculative} hallucinated future words to balance the latency/quality trade-off.

\begin{figure*}[ht!]
  \centering
  \includegraphics[width=0.7\textwidth]{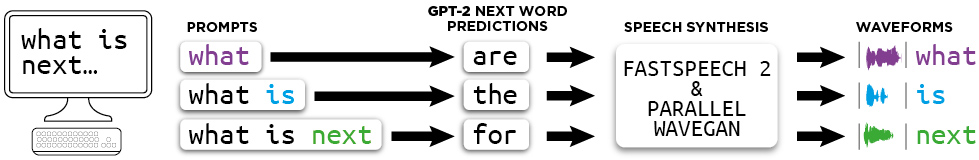}
  \caption{Utilizing language model predictions to improve incremental TTS quality while keeping limited latency.} 
  \label{fig:process}
\end{figure*}

In the present work, we propose an iTTS system that incorporates a language model to predict future lookahead.\footnote{A similar idea was proposed in a contemporary preprint paper \cite{saeki2020incremental}. 
In their work, a context encoder architecture is used.} 
Our approach (described in Figure \ref{fig:process}) predicts one word into the future. This limited lookahead was chosen so that the effects of correct and incorrect predictions could be studied.
We evaluate our system by contrasting different future word contexts: (a) unknown, (b) language model predicted, (c) randomly predicted (a control group) and (d) ground-truth (see Table \ref{table:InputTypes}).  Differences are measured at the TTS encoder level and from the generated speech signal through a listening test.

\begin{table*}[ht!]
\centering
\caption{\label{tab:InputTypes}Examples of input sequences with unknown, ground-truth, predicted and random future context. In each sequence, the word in bold is the word which is synthesized from the sequence.}
\begin{tabular}{|l|l|l|}
\hline
\small \textbf{Input Type} & \small \textbf{Lookahead}                                            & \small \textbf{Input Sequences}                         \\ \hline
\small Ground Truth        & \small \begin{tabular}[c]{@{}l@{}}Full sentence, $k = N-n$ \end{tabular} &  \small \textbf{Do you think that you could manage, Tidy?}\\ \hline
\small Unknown (future)       & \small $k=0$ word                                                          & 
\small $\mathbf{s}_{1:n+0}^{\text{GT}}$ = \textbf{Do}, Do \textbf{you}, Do you \textbf{think}, ...                                    \\[5pt] \hline
\small Ground Truth        & \small $k=1$ word                                                          &

\small $\mathbf{s}_{1:n+1}^{\text{GT}}$ = \textbf{Do} you, Do \textbf{you} think, Do you \textbf{think} that, ...                         \\[5pt] \hline
\small GPT-2 prediction    & \small $k=1$ word                                                          & 
\small $\mathbf{s}_{1:n+1}^{\text{Pred}}$ = \textbf{Do} they, Do \textbf{you} agree, Do you \textbf{think} this, ...                         \\[5pt] \hline
\small Random        & \small $k=1$ word                                                           & 
\small $\mathbf{s}_{1:n+1}^{\text{Rand}}$ = \textbf{Do} dance, Do \textbf{you} until, Do you \textbf{think} art, ...                    \\[5pt] \hline

\end{tabular}

\label{table:InputTypes}
\end{table*}

\section{Method}

\subsection{Definitions}

For each token in our corpus, we prepare different sequences which are used as input to the TTS model, FastSpeech 2 \cite{ren2020fastspeech}.
\begin{itemize}[leftmargin=*]
\itemsep0em
\item $\mathbf{x}_{1:n} = {x_1, x_2, ..., x_n}$ is the sequence of tokens up to $n$.
In the proposed iTTS system, the tokenization policy is to split the sentence on space characters, and then synthesis is triggered when a space character is encountered.

\item $k$ is the lookahead parameter (number of future tokens available when synthesizing token $x_n$).

\item $\mathbf{s}_{1:n+k} = \{x_1, x_2, ..., x_n, \hat{x}_{n+1}, ...,  \hat{x}_{n+k}\}= \{\mathbf{x}_{1:n}, \hat{\mathbf{x}}_{n+1:n+k}\}$ is the sequence used for the synthesis of token $x_n$, where for the ground-truth condition (GT) $\hat{\mathbf{x}}_{n+1:n+k}=\mathbf{x}_{n+1:n+k}$,
for the prediction condition (Pred) $\hat{\mathbf{x}}_{n+1:n+k}$ is given by the language model, and for the random condition (Rand) $\hat{\mathbf{x}}_{n+1:n+k}$ is random. The random token generation is described in Section \ref{sampling}.

\item $\mathbf{s}_{1:n}$ is the input prompt used to generate language model predictions.

\item Near the end of the sequence, we replace $n+k$ with $\min(n+k,N)$ where $N$ is the length of the full utterance.



\end{itemize}


\subsection{Models}

\textbf{Language model used for prediction.} We use the GPT-2 language model for our study. This is an auto-regressive model trained to predict the next word given a sequence of past words (causal language modeling task), based on a Transformer architecture.
The original GPT-2 \cite{radford2019language} is large (1.5B parameters) and since our intended use requires fast predictions, we opted to use a smaller version of GPT-2, called ``distilled GPT-2'' \cite{wolf-etal-2020-transformers}.\footnote{https://huggingface.co/distilgpt2} This model has been trained to produce the same output probability distribution as the original GPT-2 but using fewer layers/parameters.

\textbf{TTS model.} For TTS we select a fast and high-quality
end-to-end model: FastSpeech 2. 
The implementation we use \cite{hayashi2020espnet},\footnote{ https://github.com/espnet/espnet \label{espnet}}
trained on the LJ Speech Dataset \cite{Ito2017}, takes characters as input and converts them to phonemes. Phoneme embeddings are passed through several self-attention layers before the model makes duration, pitch and energy predictions for each phoneme. These feature predictions and the latent phoneme representations are then passed to the decoder (more self-attention layers) which produces a Mel-spectrogram.\footnote{For implementation details, see https://tinyurl.com/s7p38hcr} The Mel-spectrogram is then input into a Parallel WaveGAN vocoder \cite{9053795} (trained on full sentence inputs) for waveform generation.
This model is well suited to iTTS because (1)
it is fast which is desirable when the objective is to reduce latency (the speed is achieved by predicting all Mel-spectrogram frames in parallel), and (2) it makes explicit duration predictions for each phoneme, which makes it possible to segment words and only synthesize the word(s) of interest.

\subsection{Incremental synthesis (iTTS)} \label{incSyn}

We implement an incremental synthesis procedure where each token $x_n$ is synthesized from the input sequence $\mathbf{s}_{1:n+k}$.
Mel-spectrogram frames corresponding to individual tokens are identified using the internal duration predictions made by FastSpeech 2. Successive word-level Mel-spectrograms are input into the Parallel WaveGAN vocoder on a word-by-word basis. Resulting waveforms are concatenated together using a 1-ms crossfade to eliminate glitches (synthetic audio samples are available at https://tinyurl.com/ae4nzs).

\section{Experiments}

\subsection{Corpus and predictions} \label{sampling}

The English corpus we use for analysis consists of 1,000 sentences from LibriTTS \cite{zen2019libritts}. Sentence length ranges from $5$ to $42$ words, with a total of 16,965 tokens and 62,556 phonemes.

For each token $x_n$ in the corpus, we sampled five GPT-2 and five random next word predictions ($\hat{x}_{n+1}$). The GPT-2 predictions are constrained to the 30 most likely next words (top-30 sampling strategy). The random words were selected from a list of 1,266 of the most common words in English \cite{robyn_speer_2018_1443582}.
Importantly, we force GPT-2 predictions and random predictions to have comparable lengths in term of characters/phonemes because (1) GPT-2 tends to predict shorter words because they are more frequent, (2) in our previous study \cite{stephenson2020future}, we found that longer future words have more influence on the current token's internal representation (in a seq-to-seq model) than shorter ones, (3) otherwise, our results may be biased by the fact that the random condition simply has more future context. To control for word length in the random condition, we (1) took the word-length distribution of  GPT-2 predictions, (2) randomly sampled a word-length category from this distribution (e.g. 2-4 characters), (3) limited our most-common list to only words in this category and (4) randomly sampled a word from this list using a uniform distribution.

GPT-2 uses byte pair encoding (BPE) which breaks words down into subword units to better handle out-of-vocabulary tokens. As such, some of its predictions extend the final prompt word rather than predicting a new token (e.g. previous $\rightarrow$ previously).  To avoid such distortions to our input text, we sample until the first character in the predicted text is a space. This also prevents erroneous punctuation marks from being predicted.

\subsection{Metrics}

\textbf{FastSpeech 2 representations.} We aim at evaluating the prosody obtained in the different test conditions: no context ($k=0$), ground-truth context (GT), predicted context (Pred), random context (Rand). For this aim, we compare the pitch, duration and energy values produced in those conditions with the values produced in the \textit{reference} condition (Ref) where the full context (full sentence input) is used. In the present paper, we concentrate on the case $k=1$ (one-word lookahead). 

\begin{figure}[ht]
  \centering
  \includegraphics[width=0.42\textwidth]{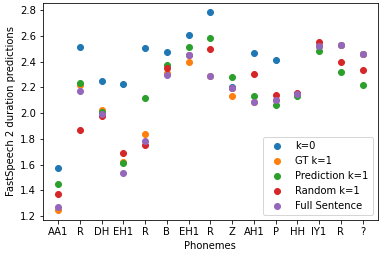}
  \caption{Duration prediction from FastSpeech2 (in number of Mel-spectrogram frames on a log scale) for each phoneme in the sentence ``Are there bears up here?'' and for the different tested prediction conditions.} 
  \label{fig:duration}
\end{figure}

As for duration and energy, they are first computed at the phoneme level, using the FastSpeech 2 internal predictions (see Figure~\ref{fig:duration} for a plot of duration values from an example sentence). A phoneme duration is defined as (the log of) the number of Mel-spectrogram frames of that phoneme. The energy is the squared magnitude of the short-time Fourier transform (STFT), averaged across all frequency bins and across the duration of the phoneme. Then the mean absolute error (MAE) is computed by averaging the absolute value of the difference of duration values obtained in each test condition and in the reference condition across all phonemes of the dataset, and the same for the energy feature. The results are reported in Table \ref{table:durEn}.

Pitch is evaluated at the sentence level.\footnote{We did not evaluate error in the internal FastSpeech 2 pitch predictions because we observed a few extreme prediction values which did not materialize in the resultant audio.} We first align the Mel-spectrograms obtained in the test and reference conditions with Dynamic Time Warping using the Librosa library  \cite{brian_mcfee-proc-scipy-2015}. Then we extract the pitch curves from the concatenated audio (see Section \ref{incSyn}) using Praat/Parselmouth \cite{praat,parselmouth} and we compute the MAE in cents between the aligned $f_0$ trajectories:
\begin{equation} \label{MAE}
\small MAE =   {1200}/{T} \sum_{t=1}^T \left\lvert  \log_2 \big(f_0^{Test}(t)/f_0^{Ref}(t)\big) \right\rvert.
\end{equation}
Then the sentence-level MAEs are averaged across all sentences of the dataset. The results are reported in Table \ref{table:Pitch}.

\begin{table}[]
\caption{MAE (and standard deviation across phonemes) between duration (resp.~energy) obtained with full context and with limited context. *unit = number of Mel-spectrogram frames on a log scale; **arbitrary unit: signal is digital, normalized and averaged.}
\begin{tabular}{|l|l|l|l|}
\hline
\textbf{\begin{tabular}[c]{@{}l@{}}\small Input type\end{tabular}}         & \textbf{\begin{tabular}[c]{@{}l@{}}\small{\# phonemes}\end{tabular}} & \textbf{\begin{tabular}[c]{@{}l@{}} \small Duration*\end{tabular}} & \textbf{\begin{tabular}[c]{@{}l@{}}\small Energy**\end{tabular}} \\ \hline \hline
\scriptsize $k=0$      & \scriptsize62,556        & \scriptsize $0.262 \pm 0.297$      & \scriptsize $0.301 \pm  0.364$                                  \\ \hline
\scriptsize GT $k=1$                                                       & \scriptsize62,556         & \scriptsize $0.077 \pm 0.133$                                     & \scriptsize $0.176 \pm 0.241$                                   \\ \hline
\begin{tabular}[c]{@{}l@{}}\scriptsize Pred $k=1$\end{tabular}               & \scriptsize 5$\times$62,556                                                                   & \scriptsize$0.135 \pm 0.198$                                     & \scriptsize $0.247 \pm 0.296$                                   \\ \hline
\begin{tabular}[c]{@{}l@{}} \scriptsize Rand $k=1$\end{tabular}                    & \scriptsize 5$\times$62,556                                                                   & \scriptsize $0.147 \pm 0.208$                                     & \scriptsize$0.260 \pm 0.304$                                   \\ \hline \hline
\begin{tabular}[c]{@{}l@{}} \scriptsize Correct pred. \end{tabular}   & \scriptsize38,274                                        & \scriptsize $0.086 \pm 0.132$                                     & \scriptsize$0.187 \pm 0.239$                                   \\ \hline
\begin{tabular}[c]{@{}l@{}} \scriptsize Incorrect pred. \end{tabular} & \scriptsize 274,506                                                                   & \scriptsize$0.142 \pm 0.205$                                     & \scriptsize$0.255 \pm 0.301$                                   \\ \hline
\end{tabular}

\label{table:durEn}
\end{table}

\begin{table}[!ht]
\begin{center}
\caption{MAE between the pitch curves obtained with the full context and with limited context.}
\label{table:Pitch}
\begin{tabular}{|l|l|l|l|}
\hline
\small \textbf{Input type}                & \textbf{\begin{tabular}[c]{@{}l@{}}\small \# sentences \end{tabular}} & \textbf{\begin{tabular}[c]{@{}l@{}}\small Pitch MAE (Cents)\end{tabular}} \\ \hline
\small $k=0$           &  \small 1,000   &  \small $203.56 \pm 45.50$                                                          \\ \hline
\small GT $k=1$    &  \small 1,000      & \small $88.57  \pm 26.33$                                                               \\ \hline
\begin{tabular}[c]{@{}l@{}} \small Pred $k=1$ \end{tabular} &  \small 5$\times$1,000          & \small $120.03        \pm 29.34$                                     \\ \hline
\small Rand $k=1$       &  \small 5$\times$1,000          & \small $123.03    \pm 30.27$                                                               \\ \hline
\end{tabular}
\end{center}

\end{table}

\begin{figure}[ht]
  \centering
  \includegraphics[width=\linewidth]{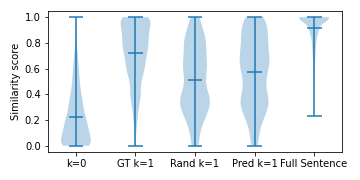}
  \caption{Violin plots of the distribution of similarity scores between signals generated with full context and signals generated with limited context for the 20 sentences in the MUSHRA test. The middle bars show the mean value.} 
  \label{fig:violin}
\end{figure}

\noindent \textbf{Perceptive test.} Finally, we evaluate the global quality using 40 native English speaking evaluators\footnote{Anonymous participants were recruited using Prolific (www.prolific.co). They were compensated at a rate slightly above the UK minimum wage.} and a MUSHRA test \cite{mushra}. We selected 20 sentences from our corpus and for each sentence, we presented the listeners with a reference audio clip (generated with the full sentence context) and then asked them to assign a similarity score to five test clips: the hidden reference (identical to the reference and used as the MUSHRA high anchor), $k=0$ (used as the low anchor), Ground-Truth $k=1$, GPT-2 prediction $k=1$ and random prediction $k=1$. We then compare the distributions of the similarity scores. The responses from four of the participants were removed because these listeners consistently failed to assign a high similarity score to the high anchor. See Figure \ref{fig:violin} for results.

\section{Discussion}

For all metrics, with regards to the mean, we see a clear ranking in the similarity to the full sentence reference: $k=0$ is farthest away, GT $k=1$ is the closest and \textit{Pred} and \textit{Rand} are in between, the former being slightly closer to full context than the latter. Statistical tests (t-test for pitch, duration and energy measures and Wilcoxon for the listening test) confirmed that \textit{Pred} and \textit{Rand} do not belong to the same distribution (p-value $<$ 0.05) and that \textit{Pred} is better by a small but significant margin. 

We notice that duration predictions for $k=0$ are almost always longer than the other conditions (Figure \ref{fig:duration}). And as in \cite{baumann_evaluating_2012}, we observe pitch drops for $k=0$ words. This is because all words are interpreted as the end of a sentence (as they are the final word in the FastSpeech 2 input, hence sentence final characteristics are predicted by the model). Both the prediction and the random conditions reduce this effect thanks to the additional padding words.

\textbf{Correct vs. Incorrect Predictions.} When we separate the correct from the incorrect GPT-2 next word predictions (see Table \ref{table:durEn}), we see that the MAE for the incorrect predictions is almost identical to the MAE for the random condition. This suggests that the improved syntactical accuracy gained from the GPT-2 predictions (the POS of the predicted token matches that of the GT next token 43.5\% of the time vs. 18.0\% for random)
does not translate into improved prosodic features.

Since we only see improvement when the exact next word is predicted, it is clear that the minor difference between GPT-2 and random is explainable by the low exact-word prediction rate.  We observe that 76\% of the GPT-2 sequences have a prediction rate lower than 10\%, and 97\% have a rate lower than 21\% (mean: GPT-2 = 6.8\%; random = 0.09\%). It is likely that as language models continue to improve \cite{brown2020language}, we will see greater gains in naturalness from the proposed method (improvements in semantic modelling will narrow the range of word choice, resulting in more frequent exact word predictions). However, these potential advances will have a fairly low ceiling if we consider human prediction abilities as the upper limit: \cite{luke2016limits} shows that approximately 5\% of context words and 20\% of function words are highly predictable by humans. Further prediction gains could be achieved if the language model was fine-tuned on the traits of a specific author \cite{delasalles:hal-02466142}; this would be an advisable step in the use case of assistive technologies for the speech impaired.

\textbf{Context Sensitivity.} Previous studies investigating the impact of lookahead have shown the contrast between $k=0$ context and different degrees of ground-truth lookahead. The setup of the present study allows us to investigate where the choice of future context modifies the output the most (i.e. where do prosodic features remain stable irrespective of the future context and where do they vary dependent on the future context). To this purpose, we calculated the range of phoneme duration and pitch feature values predicted by the TTS model in all test conditions except $k=0$. More precisely, from the 12 predicted and ground truth conditions ($5 \ \times$ Pred, $5 \  \times$ Rand, GT $k=1$ and full context), 
we take the max and min values from this set and calculate the difference. This analysis shows that a large portion of phonemes in the corpus alter only slightly when provided with different next word contexts. The pitch range does not exceed 300 cents for approximately 75\% of our samples, which falls below the Just Noticeable Difference (JND) threshold for pitch distance found by \cite{t1981differential}. The duration range is limited to a single spectrogram frame (11.75ms) for 40\% of phonemes, which, depending on the length of the phoneme, may be imperceptible to the average listener (\cite{quene2007just} found a JND of 5\%).

\begin{figure}[ht]
  \centering
  \includegraphics[width=\linewidth]{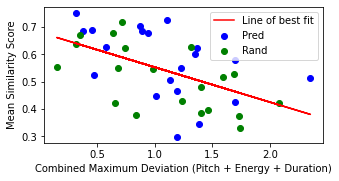}
  \caption{Each point represents a sentence (synthesized under the Pred or Rand condition) from the MUSHRA test. The x-axis shows the scaled and combined (pitch, energy, duration) maximum deviation values (deviation from the full context value) for the phonemes in the sentence. The y-axis shows the mean similarity score for (Pred,Rand) sentences to their full context counterpart, given by the MUSHRA participants. The Pearson correlation coefficient is equal to  $-0.53$.} 
  \label{fig:sentenceDev}
\end{figure}

We do however see some wide range values in the corpus which explain the large standard deviations in Table \ref{table:durEn} and the significant variability of the Pred and Rand scores in the MUSHRA test (Figure \ref{fig:sentenceDev}: the maximum deviation values in a sentence show strong correlation with the mean MUSHRA similarity scores). By examining the corpus, we notice discernible patterns in the locations of large context sensitivity. With respect to pitch, we see large variation when there is a mismatch between predicted and ground truth punctuation at the end of the next word or when there is a reporting verb (e.g. \textit{said}, \textit{exclaimed}) rather than the beginning of a new sentence following a punctuation mark.
With respect to duration, the largest variance occurs at the beginning of sentences, at punctuation marks and in function words, especially in the coordinating conjunction \textit{and}.

\section{Conclusion and Perspectives}
\vspace{-0.1cm}
The results from all metrics show that the language model predicted text does improve prosody when compared to the $k=0$ condition. Slight improvements over the random text condition are also observed.
We have seen that language model predictions are often incorrect and context mismatches can occasionally cause major distortions compared to the full context prosody. To improve our model, we could a) implement a wait policy that delays synthesis when a context sensitive word is encountered (similar to \cite{pouget_2016_interspeech_adaptive_latency_iTTS}) or b) retrain the model on both ground truth and predicted input. With this training regime, the iTTS model would find the optimal solution given contextual ambiguity (similar to \cite{pouget2015hmm}); the language model predictions, which frequently differ from the ground truth in terms of relevant contextual features (word choice, next phonemes, number of syllables, POS, and position in the prosodic phrase/utterance) would serve to make the model more robust and perhaps provide more neutral solutions for context sensitive words.



\section{Acknowledgements}

This work was funded by the Multidisciplinary Institute in Artificial Intelligence MIAI@Grenoble-Alpes (ANR-19-P3IA- 0003).

\bibliographystyle{IEEEtran}

\bibliography{custom.bib, references.bib}


\end{document}